\documentclass[11pt,twoside]{article}

\usepackage{asp2014}

\aspSuppressVolSlug
\resetcounters

\begin{document}

\title{Experimenting with Large Language Models and vector embeddings in NASA SciX}

\author{
    Sergi~Blanco-Cuaresma$^{1,2}$,
    Ioana~Ciucă$^{3,4,5}$,
    Alberto~Accomazzi$^1$,
    Michael~J.~Kurtz$^1$,
    Edwin~A.~Henneken$^1$,
    Kelly~E.~Lockhart$^1$,
    Felix~Grezes$^1$,
    Thomas~Allen$^1$,
    Golnaz~Shapurian$^1$,
    Carolyn~S.~Grant$^1$,
    Donna~M.~Thompson$^1$,
    Timothy~W.~Hostetler$^1$,
    Matthew~R.~Templeton$^1$,
    Shinyi~Chen$^1$,
    Jennifer~Koch$^1$,
    Taylor~Jacovich$^1$,
    Daniel~Chivvis$^1$,
    Fernanda~de~Macedo~Alves$^1$,
    Jean-Claude~Paquin$^1$,
    Jennifer~Bartlett$^1$,
    Mugdha~Polimera$^1$,
    and Stephanie~Jarmak$^1$.
} 

\affil{$^1$Harvard-Smithsonian Center for Astrophysics, 60 Garden Street, Cambridge, MA 02138, USA \email{sblancocuaresma@cfa.harvard.edu}}
\affil{$^2$Laboratoire de Recherche en Neuroimagerie, University Hospital (CHUV) and University of Lausanne (UNIL), Lausanne, Switzerland}
\affil{$^3$Research School of Astronomy \& Astrophysics, Australian National University, Canberra, ACT 2611, Australia}
\affil{$^4$School of Computing, Australian National University, Canberra, ACT 2601, Australia}
\affil{$^5$ARC Centre of Excellence for All Sky Astrophysics in 3 Dimensions (ASTRO 3D), Australia}

\paperauthor{Sergi~Blanco-Cuaresma}{sblancocuaresma@cfa.harvard.edu}{0000-0002-1584-0171}{Harvard-Smithsonian Center for Astrophysics}{HEAD}{Cambridge}{MA}{02138}{USA}
\paperauthor{Ioana~Ciucă}{ioana.ciuca@anu.edu.au}{0000-0001-6823-5453}{Australian National University}{}{Canberra}{ACT}{2611}{Australia}
\paperauthor{Alberto~Accomazzi}{aaccomazzi@cfa.harvard.edu}{0000-0002-4110-3511}{Harvard-Smithsonian Center for Astrophysics}{HEAD}{Cambridge}{MA}{02138}{USA}
\paperauthor{Michael~J.~Kurtz}{kurtz@cfa.harvard.edu}{0000-0002-6949-0090}{Harvard-Smithsonian Center for Astrophysics}{HEAD}{Cambridge}{MA}{02138}{USA}
\paperauthor{Edwin~A.~Henneken}{ehenneken@cfa.harvard.edu}{0000-0003-4264-2450}{Harvard-Smithsonian Center for Astrophysics}{HEAD}{Cambridge}{MA}{02138}{USA}
\paperauthor{Kelly~E.~Lockhart}{kelly.lockhart@cfa.harvard.edu}{0000-0002-8130-1440}{Harvard-Smithsonian Center for Astrophysics}{HEAD}{Cambridge}{MA}{02138}{USA}
\paperauthor{Felix~Grezes}{felix.grezes@cfa.harvard.edu}{0000-0001-8714-7774}{Harvard-Smithsonian Center for Astrophysics}{HEAD}{Cambridge}{MA}{02138}{USA}
\paperauthor{Thomas~Allen}{thomas.allen@cfa.harvard.edu}{0000-0002-5532-4809}{Harvard-Smithsonian Center for Astrophysics}{HEAD}{Cambridge}{MA}{02138}{USA}
\paperauthor{Golnaz~Shapurian}{gshapurian@cfa.harvard.edu}{0000-0001-9759-9811}{Harvard-Smithsonian Center for Astrophysics}{HEAD}{Cambridge}{MA}{02138}{USA}
\paperauthor{Carolyn~S.~Grant}{cgrant@cfa.harvard.edu}{0000-0003-4424-7366}{Harvard-Smithsonian Center for Astrophysics}{HEAD}{Cambridge}{MA}{02138}{USA}
\paperauthor{Donna~M.~Thompson}{dthompson@cfa.harvard.edu}{0000-0001-6870-2365}{Harvard-Smithsonian Center for Astrophysics}{HEAD}{Cambridge}{MA}{02138}{USA}
\paperauthor{Timothy~W.~Hostetler}{timothy.hostetler@cfa.harvard.edu}{0000-0001-9238-3667}{Harvard-Smithsonian Center for Astrophysics}{HEAD}{Cambridge}{MA}{02138}{USA}
\paperauthor{Matthew~R.~Templeton}{matthew.templeton@cfa.harvard.edu}{0000-0003-1918-0622}{Harvard-Smithsonian Center for Astrophysics}{HEAD}{Cambridge}{MA}{02138}{USA}
\paperauthor{Shinyi~Chen}{shinyi.chen@cfa.harvard.edu}{0000-0002-7641-7051}{Harvard-Smithsonian Center for Astrophysics}{HEAD}{Cambridge}{MA}{02138}{USA}
\paperauthor{Jennifer~Koch}{jennifer.koch@cfa.harvard.edu}{0000-0001-9231-8689}{Harvard-Smithsonian Center for Astrophysics}{HEAD}{Cambridge}{MA}{02138}{USA}
\paperauthor{Taylor~Jacovich}{taylor.jacovich@cfa.harvard.edu}{000-0003-0226-0343}{Harvard-Smithsonian Center for Astrophysics}{HEAD}{Cambridge}{MA}{02138}{USA}
\paperauthor{Daniel~Chivvis}{daniel.chivvis@cfa.harvard.edu}{0000-0001-6656-160X}{Harvard-Smithsonian Center for Astrophysics}{HEAD}{Cambridge}{MA}{02138}{USA}
\paperauthor{Fernanda~de~Macedo~Alves}{fernanda.de_macedo_alves@cfa.harvard.edu}{0009-0001-2353-8239}{Harvard-Smithsonian Center for Astrophysics}{HEAD}{Cambridge}{MA}{02138}{USA}
\paperauthor{Jean-Claude~Paquin}{jean-claude.paquin@cfa.harvard.edu}{0009-0001-5839-7544}{Harvard-Smithsonian Center for Astrophysics}{HEAD}{Cambridge}{MA}{02138}{USA}
\paperauthor{Jennifer~Bartlett}{jennifer.bartlett@cfa.harvard.edu}{}{Harvard-Smithsonian Center for Astrophysics}{HEAD}{Cambridge}{MA}{02138}{USA}
\paperauthor{Mugdha~Polimera}{mugdha.polimera@cfa.harvard.edu}{}{Harvard-Smithsonian Center for Astrophysics}{HEAD}{Cambridge}{MA}{02138}{USA}
\paperauthor{Stephanie~Jarmak}{stephanie.jarmak@cfa.harvard.edu}{0000-0002-6982-7191}{Harvard-Smithsonian Center for Astrophysics}{HEAD}{Cambridge}{MA}{02138}{USA}



\begin{abstract}
Open-source Large Language Models enable projects such as NASA SciX (i.e., NASA ADS) to think out of the box and try alternative approaches for information retrieval and data augmentation, while respecting data copyright and users’ privacy. However, when large language models are directly prompted with questions without any context, they are prone to hallucination. At NASA SciX we have developed an experiment where we created semantic vectors for our large collection of abstracts and full-text content, and we designed a prompt system to ask questions using contextual chunks from our system. Based on a non-systematic human evaluation, the experiment shows a lower degree of hallucination and better responses when using Retrieval Augmented Generation. Further exploration is required to design new features and data augmentation processes at NASA SciX that leverages this technology while respecting the high level of trust and quality that the project holds.
\end{abstract}



\section{Introduction}

NASA SciX\footnote{\url{https://www.scixplorer.org/}} is a digital library and search engine that provides access to a vast collection of scientific literature covering fields such as Earth Science, Planetary Science, Heliophysics and Astrophysics \citep[created by the Astrophysics Data System\footnote{\url{https://ui.adsabs.harvard.edu}};][]{2000A&AS..143...41K}. Within this project, Open-source Large Language Models (LLMs) offer the opportunity to think creatively and explore alternative methods for information retrieval and data augmentation while ensuring the protection of data copyright and users' privacy. However, when these models are directly presented with questions lacking context, they become susceptible to generating inaccurate or fictional responses (hallucinations). To address this issue, we have been experimenting with open-source (7-13 billion parameters) LLMs and our large corpus of scientific articles. This experimentation led us to build a highly customizable internal web interface and a RESTful API to easily interact with LLMs. The web interface allows users to have quick conversations with the deployed LLMs, and rapidly assess the quality of their response. On the other hand, the API enabled the NASA SciX team to develop pipelines that automatically make use of LLM capabilities for data enrichment and information extraction tasks. In this article, we will focus on our experience using LLMs to produce reasonable responses to user requests in a conversation style setup.

\section{Retrieval Augmented Generation}

One of the most common use cases of LLMs (popularized thanks to OpenAI's chatGPT) are chat sessions where the user provides instructions and the LLM generates a response back. If the user asks a question, the LLM will generate a response that is based on the data seeing during its training. For instance, if the user asks "What is the age of the universe?", most LLMs generate a response in line with "The estimated age of the universe is 13.8 billion years old.". However, LLMs will generate a response even if it was never trained with data that contained a response to the submitted questions. One recurrent question that we used in our experiments was "What is iSpec?". iSpec is a stellar spectroscopy tool \citep{2014A&A...569A.111B, 2019MNRAS.486.2075B}, and the information about this software might represent a statistically tiny fraction of the training data (or it could even be completely absent). Hence, most of the tested LLMs generate an answer that does not match reality (e.g., claiming that is a tool used in a domain that is not astrophysics). In order to minimize the risk of this kind of hallucination, we implemented the Retrieval Augmented Generation (RAG) technique, where we feed the model not only the user's question but also relevant text snippets from the scientific literature to generate a more grounded response. This way, the LLM does not rely uniquely on the knowledge acquired during training, but it can use new content that is useful to generate a response for the user.

After experimenting with multiple strategies, the prompt that worked best for most LLMs consisted of an initial system message where we provide a general contextual description of the role of the LLM (fundamentally, it is a helpful assistant interacting with a user), followed by the original user request and a set of interactions between the assistant (i.e., the LLM) and that user. These interactions were not really generated by the LLM, but we simulated them and feed them to the LLM. This way we can build an interaction that contains the textual snippets that the LLM can use to generate its final answer to the original user request. This is an example of the full prompt that we would provide to the LLM (in this case with just one snippet):

\fbox{
  \begin{minipage}{0.8\textwidth}
  \texttt{<|system|> This is a system prompt, you are a helpful assistant. Please answer truthfully and logically. If you don't know or aren't sure about something, say so clearly.</s>\\
<|user|>what is iSpec?</s>\\
<|assistant|>Can you provide me with some snippets from the literature to answer?</s>\\
<|user|>An increasing number of high-resolution stellar spectra is available today thanks to many past and ongoing extensive spectroscopic surveys. Consequently, the scientific community needs automatic...</s>\\
<|assistant|>Ok, do you have anything else?</s>\\
<|user|>This is all I found in the literature.</s>\\
<|assistant|>Awesome! I am ready to respond in a concise way.</a>\\
<|user|>Great! Go ahead!</s>\\
<|assistant|>
}
  \end{minipage}
}

The LLM receives the above prompt, and it generates a response by predicting the next token until a certain stop token is emitted, which signals the end of the response. We tested this strategy with a long list of open-source LLMs from 7 to 13 billion parameters, and based on our very subjective impression, we found Zephyr \citep{tunstall2023zephyr} to be one of the best 7 billion parameters model (as of November 2023).

\section{Finding Relevant Snippets}

The hardest part of implementing RAG is developing a system that allows to identify what are the most relevant text snippets to provide to the LLM. If the selected snippets contain misleading information or are not strictly related to the user's request, then the generated response is going to have a lower quality and it will not satisfy the user's needs. We implemented two different approaches to identify relevant text snippets.

\subsection{The traditional NASA SciX search approach}

NASA SciX search engine has been fine-tuned to serve the scientific community. It allows scientist to query the literature using extremely well curated metadata, textual matches, acronym expansion and synonyms (among other features). The corpus contains more than 20M entries, and it can be queried in an automatic way using an API. To take advantage of all this power, we decided to use a LLM to translate the user question into a request compatible with the NASA SciX search engine. This strategy allows us to retrieve the abstracts of the most relevant recent refereed articles.

To successfully translate user requests to valid NASA SciX request, we followed a similar strategy as described above. We provide the LLM with a system instruction that set the scene, describing the interaction between a user and an assistant that is an expert librarian who receive questions and response back with the best valid NASA SciX request to find answers. Then, we simulate a conversation where a user has already submitted multiple questions, and the assistant has built valid queries. Finally, we add the current user request and we let the LLM complete the dialog:

\fbox{
  \begin{minipage}{0.8\textwidth}
  \texttt{<|system|> You are an expert librarian in creating structured queries to be submitted to NASA SciX. The system accepts queries using the Apache Solr search syntax. Available search fields include "author", "abs", and "year".</s>\\
<|user|>What was written by Michael Kurtz in 2016?</s>\\
<|assistant|>((author:"Kurtz, Michael") AND (year:2016))</s>\\
<|user|>What are blackholes?</s>\\
<|assistant|>(abs:(black holes))</s>\\
<|user|>what is iSpec?</s>\\
<|assistant|>
}
  \end{minipage}
}

However, despite the detailed instruction and the provided examples, the LLM may end up generating queries that are invalid or it would add superfluous text such as "Sure! Here is your query:". To avoid this problem, we imposed that the picked tokens should comply with a predefined grammar that is compatible with the NASA SciX service. Once the natural language question to NASA SciX query translation is completed, we used the NASA SciX API to issue a search and retrieving the abstracts from the top N results. 

\subsection{The modern Semantic Search approach}

We went through all the indexed open access articles in the NASA SciX corpus, segmented the full-text content into small paragraphs, computed semantic vectors (i.e., embeddings) for each paragraph using the \textit{BAAI/bge-small-en} model \citep{bge_embedding} and ingested them into a PostgreSQL database. Then, user's requests can also be transformed into a vector and, thanks to the \textit{pg\_vector} extension, we can find what embeddings are closer, thus retrieving semantically relevant paragraphs.

\section{Conclusions}

Implementing a internal web interface and API to easily interact with open-source LLMs has helped the NASA SciX team to better understand the limits and potential use cases of these models. In particular, we have seen how following the RAG approach offers substantial benefits such as being able to list the concrete sources uses to generate a response and reduce the number of hallucinations by only relying on abstracts. Additionally, we observed that the use of full-text snippets improves the overall quality of the generated responses, making them more detailed and specific.

It is important to highlight that the results presented here are part of an internal NASA SciX experiment aimed at exploring the use of LLMs within the project. The developed web interface and services are not intended to be deployed openly to the general public. We recognize that there are costs and risks associated to openly offering a service of this characteristics. For example, currently the model utilizes one GPU, and it can only answer one request at a time. Allowing multiple parallel requests would require having at our disposal a larger number of GPUs. However, this experiment has been key to explore how NASA SciX could better serve its users by relying on LLMs for tasks such as data enrichment, classification, and knowledge extraction.

\bibliography{C407}  


\end{document}